# Domain Adaptation Broad Learning System Based on Locally Linear Embedding

Chao Yuan，Chang-E Ren

*Abstract*—Broad learning system (BLS) has been proposed for a few years. It demonstrates an effective learning capability for many classification and regression problems. However, BLS and its improved versions are mainly used to deal with unsupervised, supervised and semi-supervised learning problems in a single domain. As far as we know, a little attention is paid to the cross-domain learning ability of BLS. Therefore, we introduce BLS into the field of transfer learning and propose a novel algorithm called domain adaptation broad learning system based on locally linear embedding (DABLS-LLE). The proposed algorithm can learn a robust classification model by using a small part of labeled data from the target domain and all labeled data from the source domain. The proposed algorithm inherits the computational efficiency and learning capability of BLS. Experiments on benchmark dataset (Office-Caltech-10) verify the effectiveness of our approach. The results show that our approach can get better classification accuracy with less running time than many existing transfer learning approaches. It shows that our approach can bring a new superiority for BLS.

## I. Introduction

Broad learning system (BLS) is proposed for solving the computational complexity that may appear in some deep learning algorithms. In BLS, feature nodes are mapped randomly from input data and enhancement nodes are also mapped randomly from feature nodes, so it is a stochastic neural network. Although the weights and biases of BLS from input layer to hidden layer are generated randomly, it is rigorously proved in [3] that BLS still has good nonlinear approximation ability.

Generally speaking, BLS can approach any continuous function well in Euclidean space. BLS uses the sum of squares of training errors to evaluate the loss and uses pseudo-inverse to calculate the weights between the hidden layer and the output layer. Therefore, the rapidity of the model training of BLS can be ensured. BLS shows a comparable prediction accuracy with less training time than support vector machine (SVM), hierarchical extreme learning machine (HELM) and convolutional neural network (CNN) [2].

Many researchers have put forward several improved versions of BLS in recent years. The representative results are listed as follows. Some researchers improve the network performance by changing the structure of BLS. In details, Li et al. [4] presented a local receptive field based broad learning system (BLS-LRF). BLS-LRF retains the feature nodes and enhancement nodes of BLS, and adds a series of convolution and pooling layers in the network to extract local features. Experiments in [4] show that BLS-LRF can identify images faster and more accurately. Jia et al. [[4]] proposes a novel model to deal with multi-modal learning problems. In [5], firstly, two broad learning systems are used to extract different abstract features. Then the model obtains a representation of these features by nonlinear fusion. Finally, the model uses these features to train a classifier. Feng et al. [6] replaces the feature nodes of BLS by Takagi-Sugeno fuzzy subsystems, then the model can be used to fuzzy reasoning. Some other methods improve the performance of BLS by changing its objective function. In order to deal with various noises with different probability distribution and improve the robustness of BLS, Zheng et al. [7] changes $L_2$-norm of training error in the objective function to p-norm. In order to make the classification model be more discriminative, Jin et al. [8] adds a manifold regularization term to the objective function of BLS. In addition, some researchers apply BLS to practical problems. For example, Zhang et al. [9] proposes an algorithm based on BLS and graph theory to analyze EEG data more effectively, and it designs a graph convolution broad network to explore the deep information of graph structure data. Kong et al. [9] incorporates class-probability into BLS and proposes a semi-supervised broad learning system to classify hyperspectral images.

The improvements of BLS in [4]-[10] mainly deal with machine learning problem in single domain. That is, these methods are based on the same distribution of training data and testing data. However, the distribution of data obtained under different circumstances may be relevant but different. As an example of image classification, images in different domains may differ in perspective, background, illumination, texture and so on. Under these conditions, a model learned from the training data cannot be applied to the testing data directly. When the situation that training data and testing data may have different distribution is considered, transfer learning can be adopted.

In transfer learning, the model is usually trained by the data from the source domain, and then it is applied to the

This work was supported in part by the National Natural Science Foundation of China under Grant 61803276, Beijing Municipal Education Commission Science Plan (General Research Project, No. KM201910028004), Beijing Natural Science Foundation (4202011), Key Research Grant of Academy for Multidisciplinary Studies of CNU (JCKXYJY2019018), the National Key Research and Development Program of China under number 2019YFA0706200 and 2019YFB1703600. (Corresponding author: Chang-E Ren.)

Chang-E Ren is with the college of information engineering, Capital Normal University, Beijing, 100048, China (e-mail: dtlrce@gmail.com).
Chao Yuan is with the college of information engineering, Capital Normal University, Beijing, 100048, China (e-mail: 2181002078@cnu.edu.cn).

target domain for prediction. Adapting the regression model or classifier trained from source domain so that it can be used in target domain is often called domain adaptation problem. Many methods have been proposed in domain adaptation in recent years. Some of them, such as prediction re-weighting for domain adaptation (PRDA) [11] and the kernel mean matching (KMM) [12] focus on reducing distribution discrepancy by re-weighting the samples. Then a standard classifier learned from the re-weighted samples will get high accuracy in target domain. The standard classifier can be SVM or K-nearest neighbor. In addition, some approaches focus on learning a shared feature subspace. These approaches usually use the maximum mean discrepancy to measure the discrepancy between domains in subspace and find the invariant representations by minimizing the distribution divergence. Then a standard classifier can be implemented with the obtained domain invariant features to make predictions for the samples from the target domain. Joint distribution adaptation (JDA) [13] is a representative of this kind of methods.

Due to the strong feature extraction ability of deep learning model, many algorithms combining deep learning with domain adaptation have been proposed in recent years. Here are some typical examples. Gabourie et al. [14] uses the output space of the deep encoder to reconstruct the embedding space, and uses the sliced-Wasserstein distance to minimize the distance between the embedding distributions of the source domain and the target domain. Ganin et al. [15] combines deep transfer neural network with adversarial mechanism and proposes a domain-adversarial neural network (DANN). DANN has the advantage of selecting transferable features from different domains.

BLS is a machine learning model comparable to deep neural network. Compared with the deep neural network, BLS can train a high-accuracy model more quickly. However, the potential of BLS in domain adaptation has got a little attention. To our best knowledge, up to now, only [16] and [17] have studied BLS based domain adaptation. Although BLS has the advantage of a fast training speed and high accuracy, the obtained model in [16] does not have a good classification accuracy and robustness in the experiment. This may be caused by the fact that the method in [16] destroys the geometric structure of the data and ignores the impact of the imbalance of categories on the classification results. In [17], as the model needs to be solved by the alternating direction method of multipliers, it is time-consuming to establish the model. Therefore, how to improve the capacity of BLS for transfer learning is worthy of further study.

This paper mainly studies the domain adaptation broad learning system. We combine manifold learning theory with BLS and introduce a coefficient to reduce the impact of class imbalance on the model performance when the objective function is designed. Our research assumes that all data in the source domain have labels, while only a small part of data in the target domain have labels. The proposed algorithm is applied to benchmark dataset (Office Caltech-10) to verify its effect.

The contributions of this paper are listed in the following details.

1) In this paper, an algorithm called domain adaptation broad learning system based on locally linear embedding (DABLS-LLE) is proposed. This algorithm can deal with the problem of transfer learning in image classification. The introduction of local linear embedding theory into BLS can preserve the local geometric structure of the training data, so the proposed algorithm in this paper has good robustness and classification accuracy.

2) In order to reduce the impact of class imbalance on the classification model, a coefficient is introduced into the objective function of DABLS-LLE. The introduction of this coefficient is conducive to the classification model to fully consider the importance of the categories with less samples, and avoid over fitting of the categories with more samples. Therefore, the robustness of the model is further guaranteed.

3) Different from many deep neural networks which use gradient descent method to update weights, our proposed algorithm can use pseudo inverse to get the optimal solution directly. It ensures our algorithm more efficient in modeling.

4) As far as we know, there are a few researches in transfer learning based on BLS. A large number of experiments show that our algorithm can effectively solve the problem of transfer learning in image classification better than some existing algorithms.

The structure of this paper is as below. We briefly review BLS and LLE in Section II. In Section III, our proposed domain adaptation broad learning system based on locally linear embedding (DABLS-LLE) with its details is presented. We conduct the experiments on some benchmark datasets and the results are shown in Section IV. In Section V, we analyze the running time and classification accuracy of the algorithm. Finally, Section VI concludes the paper.

II. BASIC METHODS

A. *Broad Learning System*

Fig. 1 shows the basic network structure of BLS. BLS can be regarded as a three-layer neural network, which is mainly composed of input layer, hidden layer and output layer. Among them, the hidden layer contains two parts, which are called feature mapping nodes and enhancement mapping nodes. The basic network structure of BLS has many similarities with random vector functional-link network (RVFLNN). But different from RVFLNN, the feature mapping nodes of BLS are no longer original input data, but the data after transformation.

The detailed description of BLS is as follows. Consider a *C* classes classification problem. Suppose *N* is the number of the training data, $D$ is the number of features of each data. $x_i = (x_{i1}, x_{i2},...,x_{iD}) \in R^D$ represents an input data and its corresponding label is $y_i = (y_{i1}, y_{i2},...,y_{iC}) \in R^C$ with $i = 1,2,...N$. Then all the input data of BLS can be represented as $X = (x_1, x_2,...,x_i,...,x_N)^T \in R^{N \times D}$. $X$ stands for an input matrix composed of a series of input vectors and $Y = (y_1, y_2,...,y_i,...,y_N)^T \in R^{N \times C}$ stands for the corresponding label matrix. In Fig. 1, $M_1, M_2,..., M_n$ are the feature nodes and $E_1, E_2,..., E_m$ are the enhancement nodes. $W$ stands for the weight matrix between the hidden layer and the output layer.

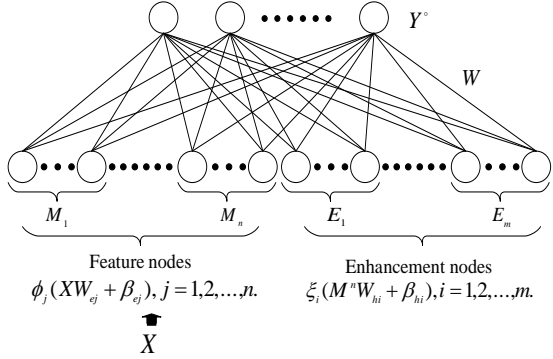

Fig.1. Structure of BLS.

From Fig. 1, we can see that $X$ is mapped to the $j$th group of feature node to get $M_j = \phi_j(XW_{ej} + \beta_{ej}) \in R^{N \times q}, j = 1,2,...,n$. In total, $n$ groups of such feature nodes are formed, which are recorded as $M_1, M_2,..., M_n$. The activation function $\phi_j$ is selected according to the actual needs, and it can be linear or nonlinear. At present, the commonly used activation functions are rectified linear unit (ReLU), sigmoid and tanh. $q$ is the number of feature nodes corresponding to each feature mapping, $W_{ej} \in R^{D \times q}$ is the weight matrix generated randomly and $\beta_{ej} \in R^{N \times q}$ is a randomly generated bias matrix. To get the sparse representation of input data, $W_{ej}$ and $\beta_{ej}$ can be finetuned by sparse auto-encoder. Then $n$ groups feature nodes are spliced together and recorded as $M^n = [M_1, M_2,...,M_n] \in R^{N \times nq}$. $M^n$ is sent to the enhancement nodes for further nonlinear transformation. The $i$th group of enhancement nodes is recorded as $E_i = \xi_i(M^n W_{hi} + \beta_{hi}) \in R^{N \times r}, i = 1,2,...,m$. $\xi_i$ is also the activation function selected according to the actual needs and it is usually a nonlinear function. $r$ is the number of nodes in each group of enhancement mapping, $W_{hi} \in R^{nq \times r}$ and $\beta_{hi} \in R^{N \times r}$ represent weight matrix and bias matrix respectively, which are generated randomly. All the enhancement nodes are spliced as a whole and recorded as $E^m = [E_1, E_2,...,E_m] \in R^{N \times mr}$. The feature node matrix $M^n$ and the enhancement node matrix $E^m$ are spliced together to form $A = [M^n | E^m] \in R^{N \times F}$ with $F = nq + mr$. The matrix $A$ is the hidden layer matrix of the network. The output of the system is $Y^\circ = AW$. $Y^\circ$ is the output matrix of the system and $W \in R^{F \times C}$ is the weight matrix from the hidden layer to the output layer. Since $W_{ej}, \beta_{ej}, W_{hi}$, and $\beta_{hi}$ can be generated randomly and remain unchanged in the training process, the only weight to be learned by BLS is $W$.

BLS optimizes the following objective function to find the optimal $W$.

$$\arg\min_W (\|Y - Y^\circ\|_2^2 + \lambda \|W\|_2^2) \quad (1)$$

In equation (1), $\|Y - Y^\circ\|_2^2$ is used to minimize the training error and $\lambda \|W\|_2^2$ is used to prevent over-fitting. $\lambda$ is a coefficient generally between 0 and 1. We can obtain the optimal $W$ by setting the derivative of equation (1) with respect to $W$ to be zero. Then the optimal $W$ can be got as

$$W = (A^T A + \lambda I)^{-1} A^T Y \quad (2)$$

where $A^T$ stands for the transpose matrix of $A$ and $I$ represents the identity matrix. When $\lambda \to 0$, $A^+ = \lim_{\lambda \to 0}(A^T A + \lambda I)^{-1} A^T$ is the pseudo inverse of matrix $A$. We have $W = A^+ Y$. This is the classical way to solve BLS. Instead of using gradient descent method to update the weights gradually, BLS uses pseudo inverse to solve $W$ directly, so the solution process of BLS is very fast.

*B. Locally Linear Embedding*

Manifold learning has attracted many researchers' attention in recent decades. The key point of manifold learning is to preserve locally invariant of data. Many researchers combine manifold learning theory with machine learning methods and propose some algorithms to improve the performance of the model. Until now there are many representative manifold learning methods, for example, locally linear embedding (LLE) [19], isometric feature mapping (ISOMAP) [19], Laplacian Eigenmap (LE) [20], marginal fisher analysis (MFA) [22], and locality preserving projections (LPP) [22]. Among them, LLE is a method to preserve local geometric property of data.

In LLE, there is a given dataset $\{x_i\}_{i=1}^N$. According to the Euclidean distance between these data, we can use K-nearest neighbor (KNN) or $\varepsilon$-ball to find the $k$ nearest neighbors of each $x_i$. The obtained neighbors can be recorded as $\{x_j, j \in Q_i\}$, where $Q_i$ is a set consists of $k$ nearest neighbors' index of each $x_i$.

The main idea of LLE is that each $x_i$ can be linearly reconstructed by its $k$ nearest neighbors. For each data

$x_i$ and its neighbor set $\{x_j, j \in Q_i\}$, LLE needs to calculate the reconstruction weight $V_{ij}$ between $x_i$ and $x_j$. If $x_j$ is not in the neighbor set $\{x_j, j \in Q_i\}$, $V_{ij} = 0$. We can minimize the following reconstruction error to calculate the weight matrix $V$.

$$e(V) = \sum_{i=1}^{N} \| x_i - \sum_{j \in Q_i} V_{ij} x_j \|_2^2 \qquad (3)$$

The weight $V_{ij}$ reflects the contribution of $x_j$ to the reconstruction of $x_i$. In order to keep the translation invariance of the data, LLE constraints $V_{ij}$ by $\sum_j V_{ij} = 1$ for all $x_i$. $V$ can be obtained by solving equation (3) under Lagrange multiplier method.

LLE requires that the low dimensional embedding $y_i$ and its $k$ nearest neighbors can reflect the reconstruction relationship of the corresponding data in the high dimensional space. That is, LLE needs to optimize the following objective function.

$$\Phi(Y) = \sum_{i=1}^{N} \| y_i - \sum_{i=1}^{k} V_{ij} y_j \|_2^2 = \sum_{i=1}^{N} \| YI_i - YV_i^T \|_2^2$$
$$= \sum_{i=1}^{N} \| Y(I_i - V_i^T) \|_2^2 \qquad (4)$$

where $I_i$ is the $i$th column of the identity matrix $I$ and $V_i^T$ is the $i$th column of $V^T$. $Y$ is the matrix reconstructed by $X$. According to the properties of matrix, we can get

$$\Phi(Y) = \| Y(I - V^T) \|_2^2 = Tr(Y^T M Y) \qquad (5)$$

where $M = (I - V)^T (I - V) \in R^{N \times N}$ is a square matrix. In equation (5), $Tr(\cdot)$ represents the trace operator of matrix.

III. DOMAIN ADAPTATION BROAD LEARNING SYSTEM BASED ON LOCALLY LINEAR EMBEDDING

A. Problem Statement

We propose domain adaptation broad learning system based on locally linear embedding (DABLS-LLE) in this section. Let $D_S$ stands for the source domain and $D_T$ stands for the target domain. This paper establishes the classification model on the following basic assumptions. First, all the samples in $D_S$ have labels. These samples can be recorded as $X_S \in R^{N_S \times D}$, $N_S$ represents the number of samples in the source domain. $Y_S \in R^{N_S \times C}$ is the corresponding label matrix of $X_S$. Second, there are a few data $X_{Tl} \in R^{N_{Tl} \times D}$ in $D_T$ which are labeled by $Y_{Tl} \in R^{N_{Tl} \times C}$. $N_{Tl}$ represents the number of labeled data in target domain. Of course, there are more unlabeled data in $D_T$ are represented by $X_{Tu}$. The above assumptions indicate that we are dealing with a semi-supervised domain adaptation problem.

The activation functions used in BLS are usually nonlinear. Nonlinear activation function is often used in machine learning models because it can improve the ability of feature extraction. However, by nonlinear transformation could destroy the local consistency of data [23]. As can be seen from Fig. 2, we take sigmoid function as an example. $u_1, u_2, u_3$ and $u_4$ are the corresponding values of $d_1, d_2, d_3$ and $d_4$ after mapping. The distance between $d_1$ and $d_2$ is smaller than that between $d_3$ and $d_4$ in the input space. However, the distance between $u_1$ and $u_2$ is much larger than that between $u_3$ and $u_4$ after transforming. Then the distance information between the data is destroyed. To solve this problem, we add a regularization term to the objective function of BLS to preserve the local geometry property of training data. Then the learned output weights can be more discriminative.

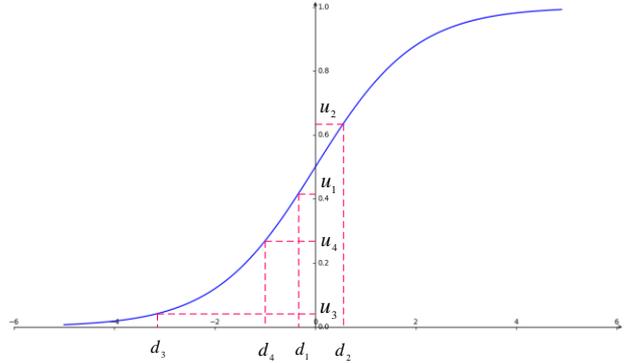

Fig. 2. The distance information is destroyed by sigmoid.

B. Proposed Method

Yang et al. [16] proposes a domain adaptation broad learning system. In view of the objective function in [16], the proposed algorithm mainly considers the approximation ability of the model, but does not fully consider the underlying geometric structure of the data. Therefore, the classification performance of the model will be slightly weakened. Simultaneously, we find that the classification problem in [16] is an unbalanced classification problem. However, the algorithm in [16] does not take into account the impact of class imbalance on classification results. This may also reduce the classification accuracy and robustness of the model. To improve the classification accuracy and robustness of machine learning model, manifold learning has been introduced in many algorithms and achieved good results [8], [24], [25]. Therefore, we introduce LLE to the objective function of BLS and propose DABLS-LLE in this paper.

The proposed DABLS-LLE algorithm contains three stages: reconstruction weights calculating, BLS feature mapping, output weights learning.

All the labeled data in the source and target domain are the training data and mark them as $X_{train} = \begin{bmatrix} X_S \\ X_{Tl} \end{bmatrix}$. First of all, we find the $k$ nearest neighbors for $X_{train}$. Let's take $X_S$ as an example. For every $x_i = (x_{i1}, x_{i2}, ..., x_{iD})$ in $X_{train}$, we have $x_i = \sum_{j \in Q_i} V_{ij} x_j$ where $Q_i$ is the set composed of the first $k$ nearest data of $x_i$ in Euclidean distance.

All $V_{ij}$ can be used to build the reconstruction weight matrix $V$ by optimizing the following objective function.

$$e(V) = \sum_{i=1}^{N_S+N_{Tl}} \| x_i - \sum_{j \in Q_i} V_{ij} x_j \|_2^2 \quad (6)$$

Next, we introduce the following regularization item into the objective function of BLS.

$$L_{LLE} = \sum_{ij} V_{ij} \| y_i^\circ - y_j^\circ \|^2 = Tr((Y_{train}^\circ)^T M Y_{train}^\circ) \quad (7)$$

In equation (7), $M = (I-V)^T(I-V)$ is a square matrix. $Y_{train}^\circ = \begin{bmatrix} Y_S^\circ \\ Y_{Tl}^\circ \end{bmatrix}$ is a matrix composed of $y_i^\circ$ which is the predicted value corresponding to labeled data.

Next, we use the procedure described in Section II to calculate the BLS features of all data in the source and target domain. Firstly, $W_{ej}$ and $\beta_{ej}$ are generated randomly and finetuned by sparse auto-encoder (SAE) for $X_{train}$. Then $M_{train}$ and $E_{train}$ can be calculated by $M_{train} = \varphi(X_{train} W_{ej} + \beta_{ej})$ and $E_{train} = \xi(M_{train}^n W_{hi} + \beta_{hi})$ with $n$ groups of feature mappings and $m$ groups of enhancement mappings. $W_{hi}$ and $\beta_{hi}$ are also randomly generated weight and bias. Thirdly, $A_{train}$ can be calculated as $A_{train} = [M_{train}^n | E_{train}^m] \in R^{N \times F}$. Note that $A_{train}$ consists of $A_S$ and $A_{Tl}$. By the same way, we can get $A_{Tu}$ by utilizing $X_{Tu}$.

Many classification problems in practical application are class imbalance problems. The characteristic of this kind of problem is that the number of samples in some categories is much larger than that in the other categories. In the process of model training, the category with less samples will be ignored, while the category with more samples will be over fitted. This may degrade the performance of the model. Therefore, we introduce a coefficient to solve this problem.

$$\tau_i = \tau_0 / N_i \quad (8)$$

where $\tau_0$ is a user-defined parameter. $N_i$ is the data's number of the class that $x_i$ belongs to. According to equation (8), when $N_i$ is large, $\tau_i$ is small. The introduction of $\tau_i$ can avoid over fitting of the model. When $N_i$ is small, $\tau_i$ is large. The introduction of $\tau_i$ can fully consider the class with less samples.

At last, we use the learned BLS features to obtain the target domain classifier $W_T$. The classifier $W_T$ should be able to correctly classify all the labeled data in the source and target domain. In addition, DABLS-LLE also take into account the local structural properties and class imbalance of data. Therefore, the objective function of DBLS-LLE is designed as follows.

$$L(W_T) = \frac{\| W_T \|_2^2}{2} + \frac{c_S \| \delta_S (Y_S - Y_S^\circ) \|_2^2}{2} + \frac{c_T \| \delta_{Tl}(Y_{Tl} - Y_{Tl}^\circ) \|_2^2}{2} + \frac{\sigma Tr((Y_{train}^\circ)^T M Y_{train}^\circ)}{2} \quad (9)$$

where $c_S$ and $c_T$ are the hyper-parameters to balance the training errors of the source domain and the target domain, respectively. $\delta_S$ and $\delta_{Tl}$ are diagonal matrices with elements $\delta_{ii} = \tau_i$ and the remaining elements are 0. $\sigma$ is the hyper-parameter of LLE regularization term.

As shown in Fig. 3, compared with BLS, the goal of DABLS-LLE is to obtain a classifier $W_T$ by all the labeled data in source domain and target domain. It extends BLS to solve cross-domain machine learning problems. To adapt the classifier $W_T$ to target domain, the regularization item of training error from a small portion of labeled data in the target domain is added to the objective function. In addition, the LLE regularization term is added to maintain the geometric structure between data. We use BLS to extract the features of all labeled data and train a classifier. Then the learned classifier can use $A_{Tu}$ to make predictions for the unlabeled data from target domain.

We bring $Y_S^\circ = A_S W_T$, $Y_{train}^\circ = A_{train} W_T$ and $Y_{Tl}^\circ = A_{Tl} W_T$ into equation (9) and the partial derivation of (9) with respect to $W_T$ is

$$\frac{\partial L(W_T)}{\partial W_T} = W_T + c_S A_S^T \delta_S^T \delta_S (A_S W_T - Y_S)$$
$$+ c_T A_{Tl}^T \delta_{Tl}^T \delta_{Tl} (A_{Tl} W_T - Y_{Tl}) + \sigma A_{train}^T M A_{train} W_T$$
(10)

by setting (10) to be zero, we can get the optimal solution

$$W_T^* = (I + c_S A_S^T \delta_S^T \delta_S A_S + c_T A_{Tl}^T \delta_{Tl}^T \delta_{Tl} A_{Tl} + \sigma A_{train}^T M A_{train})^{-1}$$
$$(c_S A_S^T \delta_S^T \delta_S Y_S + c_T A_{Tl}^T \delta_{Tl}^T \delta_{Tl} Y_{Tl})$$
(11)

Then the predictions of data from target domain can be computed by $Y_{tu}^\circ = A_{Tu} W_T^*$ (12)

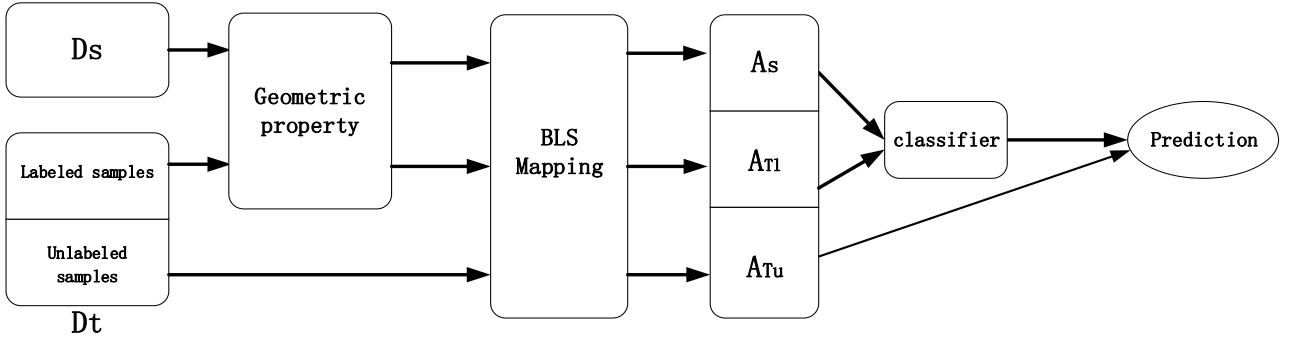

Fig. 3. Framework of DABLS-LLE.

Algorithm 1 summarizes the DABLS-LLE.

**Algorithm 1: DABLS-LLE**

Input
  Labeled source samples $\{X_s, Y_s\}$
  Labeled target samples $\{X_{Tl}, Y_{Tl}\}$
  Unlabeled target samples $\{X_{Tu}\}$
  Hyper-parameters $c_s, c_T$ and $\sigma$

Output
  The optimal target classifier $W_T^*$
  The predict output $Y_{tu}^\circ$

Procedure
**Step 1.** Compute $V_s, V_{Tl}$ using (6), compute $\delta_s, \delta_{Tl}$ using (8)
**Step 2.**
For $j \leq n$
  randomly initialize $W_{ej}, \beta_{ej}$
  compute $M_{Sj}$, $M_{Tlj}$ and $M_{Tuj}$
End
**Step 3.** Set $M_S^n = [M_{S1},...,M_{Sn}]$, $M_{Tl}^n = [M_{Tl1},...,M_{Tln}]$ and $M_{Tu}^n = [M_{Tu1},...,M_{Tun}]$
**Step 4.**
For $i \leq m$
  randomly initialize $W_{hi}, \beta_{hi}$
  compute $E_{Si}$, $E_{Tli}$ and $E_{Tui}$
End
**Step 5.** Set $E_S^m = [E_{S1},...,E_{Sm}]$, $E_{Tl}^m = [E_{Tl1},...,E_{Tlm}]$ and $E_{Tu}^m = [E_{Tu1},...,E_{Tum}]$
**Step 6.** Set $A_S = [M_S^n \mid E_S^m]$, $A_{Tl} = [M_{Tl}^n \mid E_{Tl}^m]$ and $A_{Tu} = [M_{Tu}^n \mid E_{Tu}^m]$
**Step 7.** Compute $W_T^*$ using (11), compute $Y_{tu}^\circ$ using (12)
**Step 8.** Return the weights $W_T^*$ and the predictions $Y_{tu}^\circ$.

## IV. EXPERIMENTS

In this section, we verify the effectiveness and superiority of our proposed approaches on Office+Caltech-10 dataset and compare the experimental results with other state-of-the-art transfer learning algorithms. The software and hardware environment of the experiments are as follows: Windows10, Python3.6.5, Intel-i7 2.6GHz CPU, 16G memory.

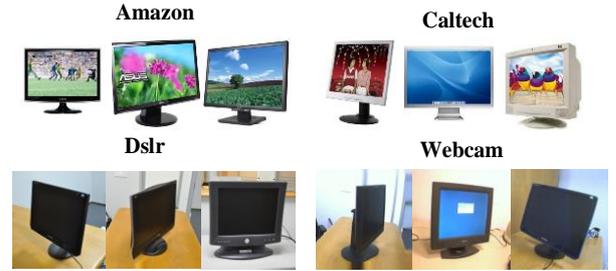

Fig. 4. Some images from Office+Caltech-10 dataset.

### A. Description of experimental data

The Office+Caltech-10 dataset is composed of Amazon (A), Caltech (C), Dslr (D) and Webcam (W). These four parts can be regarded as different domains. Fig. 4 shows some experimental samples with images. The images in Office+Caltech-10 belong to 10 categories, such as bicycle, backpack, monitor, calculator, etc. We describe the number of samples and the number of attributes in Table I. In our experiments, we select two different domains as the source domain and the target domain in turn, such as A→C, A→D, A→W, C→A..., W→A. So we implement a total of 12 cross domain learning tasks. We use one-hot to code the labels in this paper. That is to say, $y_{ij} = 1$ if $x_i$ belongs to the $j$th class and $y_{ij} = 0$ otherwise. For example, $y_i = (0,1,0,...0) \in R^C$ if $x_i$ belongs to class 2.

TABLE I    DESCRIPTION OF OFFICE+CALTECH-10 DATASET

| Datasets | Samples | Classes | Attributes | Datasets | Samples | Classes | Attributes |
|---|---|---|---|---|---|---|---|
| Amazon(A) | 958 | 10 | 800 | Caltech(C) | 1123 | 10 | 800 |
| Dslr(D) | 157 | 10 | 800 | Webcam(W) | 295 | 10 | 800 |

## B. Implementation details

In order to make a comparative experiment, we choose three standard classifiers without transfer learning, including broad learning system (BLS) extreme learning machine (ELM) [25] and SVM [26]. For these algorithms which do not have the ability of transfer learning, we train the classifier on the source domain and use this classifier to predict the samples in the target domain directly. The selected domain adaptation approaches include transfer component analysis (TCA) [27], joint distribution adaptation (JDA) [13], domain adaptation ELM (including source domain adaptation ELM (DAELM-S) and target domain adaptation ELM (DAELM-T)) [29], [30]. In addition, we also choose BLS based domain adaptation algorithms for comparative experiments. These algorithms include BLS based source domain adaptation (BLS-SDA) and BLS based target domain adaptation (BLS-TDA) [16]. Since TCA and JDA need to use standard classifier to classify the new feature representations of input data, we select SVM to complete model training task. As DAELM-S, DAELM-T in [29], BLS-SDA, BLS-TDA in [16] and our proposed algorithm are semi-supervised approaches, we select about 10% of the data in the target domain as labeled data and the remaining 90% as unlabeled data.

In the proposed approaches, there are two kinds of hyper-parameters need to be adjusted manually. One is the hyper-parameters related to the network structure of BLS: the feature dimensions in each mapped feature $q$, the number of mapped features $n$, the neurons in each enhancement mapping $m$ and the number of enhancement nodes $r$. For the convenience of experiment, the value of $m$ is set to be 1 in all experiments. The other is the hyper-parameters related to the proposed algorithm: $c_S, c_T$ and $\sigma$ in DABLS-LLE. We use the grid search method to determine the optimal values of all the above hyper-parameters, the search scopes of these hyper-parameters are shown in Table II. The best experimental results are recorded for comparison. We set the hyper-parameters of other algorithms for comparative experiments according to the methods in the corresponding references.

TABLE II    SEARCH SCOPES OF HYPER-PARAMETERS

| hyper-parameters | search scope | hyper-parameters | search scope |
|---|---|---|---|
| $n$ | {10, 20, 30, ..., 100} | $q$ | {10, 20, ..., 50} |
| $r$ | {200, 400, 600, ...,1000} | $c_S$ | {$10^{-5}, 10^{-4}, ..., 10^5$} |
| $c_T$ | {$10^{-5}, 10^{-4}, ..., 10^5$} | $\sigma$ | {$10^{-2}, 10^{-1}, ..., 10^2$} |

## C. Experiments results

We choose classification accuracy and running time as the measurement standard. In our experiments, the values of each parameter are set as follows: $q=10, n=20, r=400, c_S=10^3, c_T=10, \sigma=10^{-1}$. The experimental results of various algorithms are shown in Table III. Our proposed approach outperforms the other approaches on most of the tasks. The average accuracy of DABLS-LLE on the 12 tasks is 55.98%. Compared with BLS, the classification accuracy of our proposed method is improved by 13.12%, which indicates that DABLS-LLE can effectively improve the transferring capacity of BLS. Furthermore, compared with BLS-SDA, BLS-TDA, DAELM-S and DAELM-T, our method also achieves competitive results on Office+Caltech-10. Therefore, we make conclusion that our proposed approach can be regarded as an commendable domain adaptation algorithm.

TABLE III    COMPARISON OF CLASSIFICATION ACCURACY (%) ON OFFICE+CALTECH-10

| Method/Task | BLS | ELM | SVM | TCA | JDA | BLS-SDA | BLS-TDA | DAELM-S | DAELM-T | DABLS-LLE |
|---|---|---|---|---|---|---|---|---|---|---|
| A→C | 42.20 | 30.10 | 42.21 | 35.52 | 37.40 | 42.86 | 43.08 | 41.85 | 42.62 | 54.42 |
| A→D | 31.84 | 27.38 | 36.31 | 29.93 | 28.03 | 38.30 | 56.58 | 41.25 | 42.05 | 55.48 |
| A→W | 31.86 | 27.46 | 26.44 | 28.81 | 25.08 | 33.96 | 55.32 | 39.63 | 39.33 | 55.41 |
| C→A | 47.28 | 33.92 | 49.89 | 40.29 | 39.77 | 52.20 | 54.44 | 54.33 | 53.93 | 50.82 |
| C→D | 43.94 | 31.84 | 38.22 | 32.48 | 36.31 | 34.75 | 50.45 | 43.29 | 43.65 | 51.33 |
| C→W | 31.52 | 20.00 | 27.80 | 30.51 | 25.08 | 38.49 | 53.36 | 42.33 | 42.84 | 55.24 |
| D→A | 34.02 | 28.58 | 25.89 | 30.27 | 32.46 | 48.26 | 33.60 | 40.11 | 40.58 | 48.55 |
| D→C | 29.02 | 24.84 | 22.97 | 30.19 | 29.03 | 38.97 | 40.99 | 36.21 | 38.68 | 45.39 |
| D→W | 75.59 | 61.01 | 62.03 | 64.75 | 75.93 | 80.00 | 82.38 | 80.02 | 80.63 | 83.28 |
| W→A | 34.44 | 23.80 | 25.89 | 30.69 | 32.67 | 43.27 | 40.22 | 39.36 | 40.06 | 44.35 |
| W→C | 30.54 | 23.98 | 21.91 | 25.29 | 26.36 | 38.87 | 41.28 | 38.17 | 38.69 | 43.41 |
| W→D | 82.16 | 59.87 | 78.34 | 71.97 | 81.53 | 80.14 | 83.10 | 72.46 | 72.60 | 83.11 |
| Average | 42.86 | 31.06 | 38.15 | 37.56 | 39.14 | 47.27 | 53.02 | 47.42 | 47.97 | 55.98 |

Table IV summarizes the algorithm running time of the different approaches for different pairings of the source and target domains. From table IV, we can see that although the running time of DABLS-LLE is not the shortest, it is nearly 11 times shorter compared with the average running time of TCA and 24 times shorter compared with JDA. Our proposed approach can achieve higher classification accuracy, even though the running time of DABLS-LLE is a little longer than that of SVM,

ELM and BLS, which is worth the time balance.

TABLE IV  COMPARISON OF ALGORITHM RUNNING TIME (s) ON OFFICE+CALTECH-10

| Method/Task | SVM | ELM | BLS | TCA | JDA | DAELM-S | DAELM-T | BLS-SDA | BLS-TDA | DABLS-LLE |
|---|---|---|---|---|---|---|---|---|---|---|
| A→C | 2.23 | 0.97 | 0.27 | 86.57 | 71.71 | 2.37 | 2.48 | 2.98 | 3.02 | 3.67 |
| A→D | 1.40 | 0.53 | 0.24 | 14.32 | 62.62 | 2.23 | 2.35 | 2.36 | 3.04 | 2.82 |
| A→W | 1.52 | 0.60 | 0.22 | 20.53 | 64.56 | 2.34 | 2.45 | 2.11 | 3.21 | 2.58 |
| C→A | 2.84 | 1.02 | 0.44 | 84.83 | 72.01 | 2.02 | 2.25 | 2.65 | 3.11 | 3.38 |
| C→D | 2.01 | 0.61 | 0.25 | 20.89 | 64.41 | 2.21 | 2.29 | 2.15 | 2.96 | 2.79 |
| C→W | 2.15 | 0.86 | 0.23 | 26.86 | 66.90 | 2.36 | 2.42 | 2.57 | 2.89 | 2.77 |
| D→A | 0.25 | 0.51 | 0.41 | 14.08 | 62.81 | 2.21 | 2.38 | 2.34 | 2.87 | 2.17 |
| D→C | 0.29 | 0.75 | 0.30 | 21.12 | 64.95 | 2.45 | 2.39 | 2.22 | 2.64 | 2.42 |
| D→W | 0.11 | 0.27 | 0.19 | 0.98 | 42.21 | 2.27 | 2.47 | 1.72 | 2.42 | 1.85 |
| W→A | 0.46 | 0.71 | 0.39 | 20.18 | 65.02 | 2.31 | 2.42 | 2.21 | 2.23 | 2.61 |
| W→C | 0.51 | 0.71 | 0.31 | 26.99 | 67.22 | 2.41 | 2.48 | 2.13 | 2.34 | 2.51 |
| W→D | 0.19 | 0.26 | 0.18 | 0.99 | 42.37 | 2.06 | 2.21 | 1.83 | 2.18 | 2.10 |
| Average | 1.16 | 0.65 | 0.29 | 28.59 | 62.23 | 2.27 | 2.38 | 2.27 | 2.74 | 2.64 |

TABLE V  ACCURACY(%) COMPARISON OF ALGORITHM WITH DIFFERENT PROPORTION OF LABELED SAMPLES IN TARGET DOMAIN

| Method | BLS-SDA | | BLS-TDA | | DAELM-S | | DAELM-T | | BLS-LLE | |
|---|---|---|---|---|---|---|---|---|---|---|
| Proportion\Task | A→C | C→A | A→C | C→A | A→C | C→A | A→C | C→A | A→C | C→A |
| 10% | 42.86 | 52.20 | 43.08 | 54.44 | 41.85 | 54.33 | 42.62 | 53.93 | 54.42 | 50.82 |
| 20% | 52.13 | 55.18 | 53.26 | 54.36 | 42.51 | 55.07 | 43.74 | 55.09 | 54.66 | 60.05 |
| 30% | 51.98 | 58.15 | 53.21 | 56.29 | 46.17 | 57.96 | 46.29 | 56.11 | 53.76 | 63.93 |
| 40% | 52.92 | 61.02 | 53.38 | 62.03 | 48.16 | 59.08 | 49.05 | 60.01 | 54.52 | 65.72 |
| 50% | 55.32 | 65.51 | 54.94 | 66.81 | 53.19 | 63.22 | 53.28 | 62.25 | 56.08 | 69.52 |
| Average | 51.04 | 58.41 | 51.57 | 58.79 | 46.38 | 57.93 | 47.00 | 57.49 | 54.69 | 62.01 |

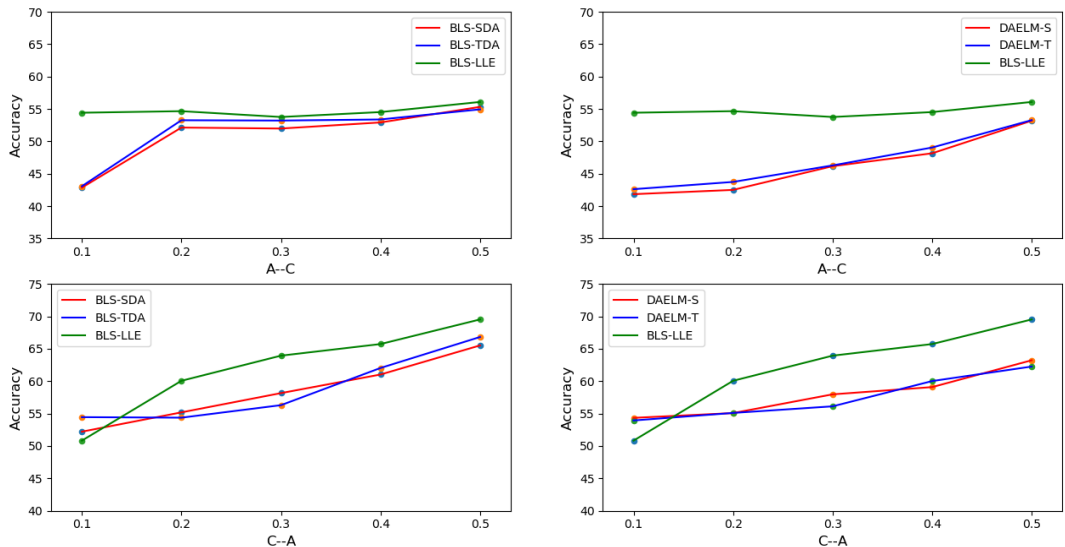

Fig. 7. Classification accuracy with different number of labeled samples in target domain.

Since we are discussing the semi-supervised transfer learning problem, in order to study the effect of the proportion of labeled samples in the target domain on the algorithm results, we carried out the transfer learning experiments on A→C and C→A with 10%, 20%, 30%, 40% and 50% labeled data in the target domain, and the experimental results are recorded in Table V. The hyper-parameters in our proposed algorithm are set as follows: $q=10, n=20, r=400, c_S=10^3, c_T=10, \sigma=10^{-1}$.
The other selected semi-supervised algorithms are BLS-SDA, BLS-TDA, DAELM-S, and DAELM-T.

To show the experimental results more intuitively, we take the proportion of labeled samples in target domain as abscissa and the classification accuracy as the ordinate to get the line chart in Fig. 7. It can be found from table V and Fig. 7 that when the number of labeled samples in the target domain increases gradually, the transfer learning results of the all algorithms are getting better and better. This shows that the more labeled samples in the target domain, the more knowledge about the target domain can be learned by the trained machine learning model, which makes the model more discriminative to the target domain. The average accuracy of BLS-SDA and BLS-TDA on A→C is about 51.04% and 51.57%. However, the accuracy of DABLS-LLE on A→C can reach 54.69%, which is about 3% higher than that of BLS-SDA and BLS-TDA. When observing the accuracy on C→A, we can get the same conclusion that the accuracy of DABLS-LLE is about 3.5% higher than that of BLS-SDA and BLS-TDA. BLS-LLE obtains better transfer learning result than BLS-SDA and BLS-TDA, which proves once again the necessity of considering local geometry of training data. Because the feature extraction process of DAELM-S, and DAELM-T is relatively simple, and they do not consider the manifold regularization term, the average accuracy of the two models is lower than that of the BLS based algorithm (including BLS-SDA, BLS-TDA, and DABLS-LLE).

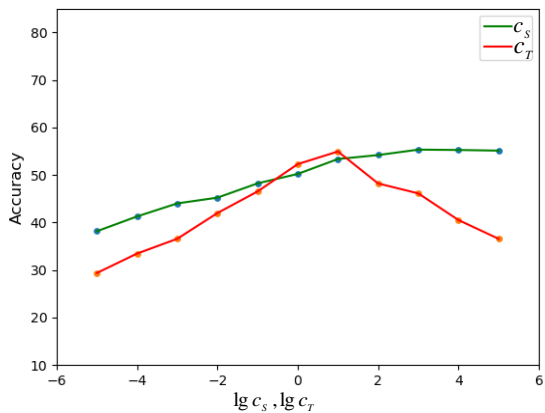

Fig. 5.  Influence of regularization parameters of source domain and target domain on experimental results.

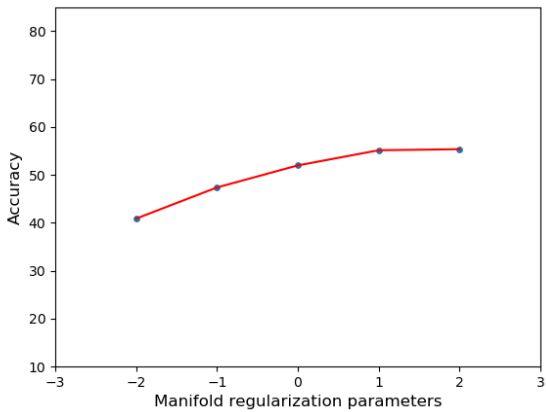

Fig. 6.  Influence of manifold regularization parameters on experimental results.

There are three hyper-parameters closely related to DABLS-LLE, namely $c_S, c_T$ and $\sigma$. We conduct experiments on A→C dataset to analyze the influence of these parameters on the experimental results. As shown in Fig. 5, we take the logarithm of $c_S$ and $c_T$ as abscissa and the classification accuracy as the ordinate. It can be seen that when the value of $c_S$ gradually increases, the classification accuracy of the model also increases. But for $c_T$, with the increase of $c_T$, the classification accuracy of the model increases first and then decreases. Referring to Fig. 6, the effect of manifold regularization parameter $\sigma$ on the experimental results is similar to $c_T$. This shows that we can find the best setting.

## V. ALGORITHM ANALYSIS

In this section, we explore more properties about the proposed approach.

1) *Accuracy.* Compared with SVM, ELM and BLS, transfer learning algorithm (including TCA, JDA, DAELM-S, DAELM-T, BLS-SDA, BLS-TDA and DABLS-LLE) get higher classification accuracy because they consider the relationship between source domain and target domain. Compared with other transfer learning algorithms, including BLS-SDA, BLS-TDA, DAELM-S and DAELM-T, the proposed DABLS-LLE can get higher classification accuracy because of considering the local geometry structure between data. In some transfer learning tasks, the accuracy of our proposed method can reach about 83.28%, which may be caused by the large similarity between these datasets.

2) *Running time.* TCA and JDA need to map the source domain data and the target domain data to the same feature space, which greatly increases the running time of the algorithm, so the running time of these two algorithms are longer than the others. Because the transfer learning algorithm based on BLS and ELM can directly obtain the optimal solution by ridge regression, BLS-SDA, BLS-TDA, DAELM-S, DAELM-T and DABLS-LLE can run rapidly. DABLS-LLE need to construct the reconstruction matrix to reflect the geometry structure between data, which lead to the running time is longer than the other transfer learning approaches based on BLS and ELM, but the time does not increase much.

3) *Hyper-parameters.* Due to the large number of labels in the source domain, setting the regularization parameter $c_S$ larger can better play the auxiliary role of the source domain to help the model get better classification results in the target domain. However, the experimental results show that $c_T$ should be set as a suitable value near 1, because there are fewer labels in the target domain. If $c_T$ is too large, it will lead to the model to over fit. As for the manifold regularization parameter $\sigma$, it should not be too large because manifold learning fully maintains the local linear relationship of the source domain data, but these linear relationships do not necessarily conform to the data distribution of the target domain.

## VI. CONCLUSION

We extend BLS to domain adaptation problem in this paper and propose an approach to transfer knowledge from source domain to target domain, namely DABLS-LLE. By combining domain adaptation and Locally Linear Embedding in the objective function, our proposed transfer learning algorithm can better reflect the local geometric properties of the image, so the proposed algorithm can improve the accuracy of image classification. The optimal solution of the objective function can be easily obtained by ridge regression theory, which ensures the efficiency of the model. The experiments on Office+Caltech-10 show that our proposed approaches can improve transferring ability of BLS.